\documentclass[11pt]{article}

\usepackage[english]{babel}
\usepackage[letterpaper,top=2cm,bottom=2cm,left=3cm,right=3cm]{geometry}
\usepackage{graphicx}
\usepackage{amsmath, amssymb}
\usepackage{hyperref}
\usepackage{authblk}
\usepackage{cite}
\usepackage[font=small,labelfont=bf]{caption}
\usepackage{makecell} 
\usepackage{booktabs} 
\usepackage{float}

\title{OmniMRI: A Unified Vision--Language Foundation Model for Generalist MRI Interpretation}

\author[1,2]{Xingxin He, PhD}
\author[1,3]{Aurora Rofena, BS}
\author[1,2]{Ruimin Feng, PhD}
\author[1]{Haozhe Liao, BS}
\author[2]{Zhaoye Zhou, PhD}
\author[1,2]{Albert Jang, PhD}
\author[*,1,2]{Fang Liu, PhD}

\affil[1]{Athinoula A. Martinos Center for Biomedical Imaging, Harvard Medical School, Boston, MA, United States}
\affil[2]{Department of Radiology, Massachusetts General Hospital, Boston, MA, United States}
\affil[3]{Unit of Computer Systems and Bioinformatics, Department of Engineering, University Campus Bio-Medico of Rome, Rome, Italy}

\affil[*]{Corresponding author: \texttt{fliu12@mgh.harvard.edu}}

\date{}

\begin{document}

\maketitle

\begin{abstract}
Magnetic Resonance Imaging (MRI) is indispensable in clinical practice but remains constrained by fragmented, multi-stage workflows encompassing acquisition, reconstruction, segmentation, detection, diagnosis, and reporting. While deep learning has achieved progress in individual tasks, existing approaches are often anatomy- or application-specific and lack generalizability across diverse clinical settings. Moreover, current pipelines rarely integrate imaging data with complementary language information that radiologists rely on in routine practice. Here, we introduce OmniMRI, a unified vision-language foundation model designed to generalize across the entire MRI workflow. OmniMRI is trained on a large-scale, heterogeneous corpus curated from 60 public datasets, over 220,000 MRI volumes and 19 million MRI slices, incorporating image-only data, paired vision-text data, and instruction-response data. Its multi-stage training paradigm, comprising self-supervised vision pretraining, vision-language alignment, multimodal pretraining, and multi-task instruction tuning, progressively equips the model with transferable visual representations, cross-modal reasoning, and robust instruction-following capabilities. Qualitative results demonstrate OmniMRI’s ability to perform diverse tasks within a single architecture, including MRI reconstruction, anatomical and pathological segmentation, abnormality detection, diagnostic suggestion, and radiology report generation. These findings highlight OmniMRI’s potential to consolidate fragmented pipelines into a scalable, generalist framework, paving the way toward foundation models that unify imaging and clinical language for comprehensive, end-to-end MRI interpretation.
\end{abstract}

\section{Introduction}
Magnetic Resonance Imaging (MRI) remains one of the most powerful non-invasive tools for diagnostic radiology, offering unparalleled soft tissue contrast and high-resolution anatomical detail across a wide range of neurological, musculoskeletal, and oncological conditions\cite{mcrobbie2017mri}. However, despite its clinical value, MRI workflows are inherently complex and resource-intensive, spanning time-consuming acquisition and image reconstruction\cite{lustig2007sparse}, labor-intensive anatomical landmark measurement\cite{heimann2009statistical}, and image interpretation and radiological report writing that heavily rely on the expertise of radiologists\cite{langlotz2006radlex,kahn2009toward}. These multi-stage pipelines contribute to substantial costs, workflow inefficiencies, and variability in diagnostic outcomes, especially amid increasing imaging volumes and a persistent shortage of trained radiologists.

Deep learning (DL) methods\cite{lecun2015deep} have made significant progress in automating specific steps of the MRI pipeline, such as reconstruction from undersampled k-space data\cite{safari2026advancing,knoll2020deep,zeng2021review}, anatomical or pathological segmentation\cite{dorfner2025review,akkus2017deep,asgari2021deep}, abnormality detection\cite{bouhafra2025deep,shoeibi2021applications}, and disease classification or prediction\cite{ahsan2022machine,mazurowski2019deep,adam2023deep}. However, most existing models are narrowly trained for a single task, anatomy, or contrast setting, which limits their generalizability across diverse patient populations, scanner types, acquisition protocols, and pathologies. Building separate models for each step not only increases engineering burden and system complexity but also prevents the exploitation of shared representations and inter-task consistency. In addition, conventional supervised pipelines often fail to incorporate linguistic information that is central to clinical practice. Language conveys patient symptoms and medical history, provides nuanced context beyond imaging, and forms the basis of radiological documentation, clinical guidelines, and medical knowledge bases, underscoring its indispensable role in medical decision-making.

Foundation models (FMs)\cite{bommasani2021opportunities} have emerged as a promising paradigm to overcome these limitations. By pretraining on large-scale, heterogeneous, and largely unlabeled datasets, FMs learn transferable representations that can be adapted to new scenarios with minimal (few-shot) or even no supervision (zero-shot)\cite{he2020momentum,chen2020simple}, thereby improving generalization across data domains and tasks across computer vision (e.g., DINO\cite{caron2021emerging}, MAE\cite{he2022masked}, iBOT\cite{zhou2021ibot}) or natural language processing (e.g., BERT\cite{24_devlin2019bert}, GPT\cite{25_brown2020language}, LLaMA\cite{26_touvron2023llama}). Furthermore, Transformer\cite{27_ashish2017attention}-based architectures are inherently well-suited for multimodal integration, enabling the mapping of visual patches, textual tokens, and metadata into a shared representation space. This capability facilitates vision–language reasoning, as demonstrated by recent advances such as CLIP\cite{28_radford2021learning}, BLIP\cite{29_pmlr-v162-li22n}, Flamingo\cite{30_alayrac_flamingo_2022}, LLaVA\cite{31_NEURIPS2023_6dcf277e}. Motivated by these advances, a few recent reports, such as BrainSegFounder\cite{32_COX2024103301}, MedSAM\cite{33_10635844}, FATE-SAM\cite{34_he2025fewshotadaptationtrainingfreefoundation}, and VISTA3D\cite{35_He_2025_CVPR} have shown the significant potential of applying the FMs in improving MRI segmentation and visual processing. 

\begin{figure}[htbp!]
    \centering
    \includegraphics[width=0.95\textwidth]{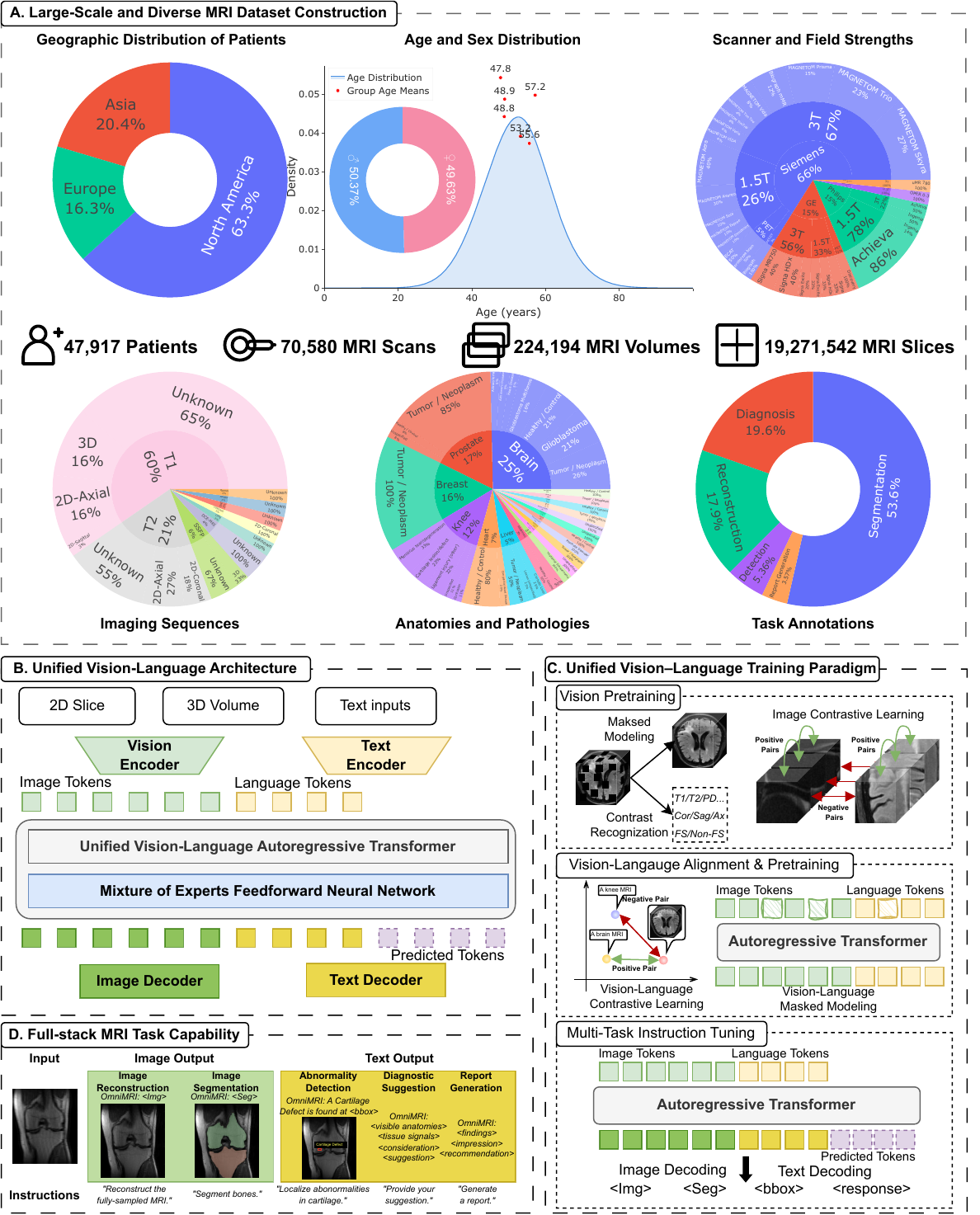}
    \caption{Overview of OmniMRI framework. (A) Large-scale and diverse MRI dataset construction. The training corpus integrates heterogeneous public datasets, covering diverse patient demographics (age, sex, geography), scanner vendors and field strengths, acquisition protocols and sequence types, anatomical regions, and task-specific annotations. (B) Unified vision–language architecture. MRI inputs (2D slices or 3D volumes) and textual inputs (metadata, prompts, or reports) are processed by modality-specific encoders, fused into a shared autoregressive Transformer backbone with a Mixture-of-Experts feedforward module, and decoded into either images or text depending on the task. (C) Unified training paradigm. The model is trained through a multi-stage process: vision pretraining on image-only data, vision–language alignment and multimodal pretraining with paired image–te4xt data, and multi-task instruction tuning using instruction–response data. (D) Full-stack MRI task capability. OmniMRI supports a wide range of downstream tasks, including image reconstruction, segmentation, abnormality detection, diagnostic suggestion, and radiology report generation, within a single unified framework.}
    \label{fig:1}
\end{figure}

In this work, we propose OmniMRI, a unified MRI foundation model designed to generalize across the entire MRI workflow (Figure \ref{fig:1}), built upon core principles: 1) Large-scale and diverse data construction (Figure \ref{fig:1}A). We collected and curated MRI datasets from 60 data sources spanning diverse geometric populations, institutions, vendors, field strengths, anatomies, and pathologies, ensuring broad coverage of clinical variability. 2) Unified vision-language architecture (Figure \ref{fig:1}B). OmniMRI processes both visual and textual inputs, including 2D slices, 3D volumes, metadata, and natural language instructions, through a shared autoregressive Transformer backbone with multimodal self-attention and a mixture-of-experts feedforward network. This design supports multi-modal integration and flexible task adaptation without the need for separate models. 3) Unified vision–language training paradigm (Figure \ref{fig:1}C). A multi-stage pipeline combines Vision pretraining, vision–language modeling, and multi-task instruction tuning, enabling OmniMRI to follow natural language prompts and seamlessly switch across pixel-level and semantic-level tasks. Guided by task instructions, OmniMRI performs image reconstruction, segmentation, abnormality detection, diagnostic suggestion, and radiology report generation within a single model (Figure \ref{fig:1}D). This unified framework consolidates task-specific pipelines into a generalist solution, enabling zero-shot transfer across contrasts, anatomies, and tasks, and integrating full-stack MRI tasks into a single scalable system.

\section{Training Dataset Construction}
A critical prerequisite for training a generalist foundation model is access to a large-scale, diverse, and well-structured dataset that captures the full complexity of clinical MRI workflows. Here, we present the collection and construction of the OmniMRI training corpus, designed to enable unified learning across full-stack MRI tasks. The data construction process consists of two main steps. First, we identify and integrate heterogeneous sources to ensure abundance and diversity of MRI data (§2.1). Second, we organize the collected data into three categories tailored to different training stages (§2.2): (1) image-only data for vision pretraining; (2) paired vision–text data for vision-language alignment and pretraining; and (3) instruction-response data for multi-task instruction tuning.

\subsection{Data Source}

The OmniMRI dataset was constructed by curating and harmonizing data from 60 publicly available MRI datasets, ensuring both scale and diversity (Figure \ref{fig:1}A). In total, it encompasses 47,917 patients, 70,580 scans, 224,194 volumes, and over 19 million slices, spanning multiple continents (North America 63.3\%, Asia 20.4\%, Europe 16.3\%) with a wide age range (mean ages 47–57 years, ranging from 1-87 years) and balanced sex ratio (male 50.37\%, female 49.63\%). Acquisitions were collected from major vendors (Siemens, GE, Philips) across multiple scanner models and field strengths (1.5T, 3T), with comprehensive sequence coverage including T1-, T2-, FLAIR-, PD-, and diffusion-weighted scans, in 2D or 3D formats. Anatomical diversity spans the brain (25\%), breast (16\%), knee (12\%), prostate (7\%), and other organ systems, covering a broad spectrum of pathologies. 

Crucially, the dataset integrates extensive supervision for downstream applications, with annotations distributed across segmentation (53.6\%), diagnosis (19.6\%), reconstruction (17.9\%), and detection (5.5\%). This large-scale, heterogeneous, and multi-annotated resource provides the foundation needed to develop a unified generalist model for full-stack MRI tasks.

\subsection{Data Construction}

To support unified vision–language pretraining and multi-task adaptation, we constructed a large-scale multimodal MRI dataset specifically aligned with the training paradigm. The process integrates heterogeneous sources, harmonizes diverse data formats, and provides annotations (where available) for multi-task supervision. In particular, we designed three complementary datasets, each tailored to a distinct stage of model training: 1) Image-only data for vision pretraining. 2) Paired vision-text data for vision–language alignment and pretraining. 3) Instruction-response data for multi-task instruction tuning.

\subsubsection{Image Only Data}

Image-only data refers to MRI volumes not related to any specific annotations, which are critical for self-supervised vision pretraining. This type of data enables the model to learn anatomy- and contrast-aware representations independent of manual labels, providing a strong foundation for visual understanding. To construct this corpus, we curated MRI scans from DICOM files as well as raw k-space data.

\subsubsection{Paired Vision-Text Data}

\begin{figure}[t]
    \centering
    \includegraphics[width=0.95\textwidth]{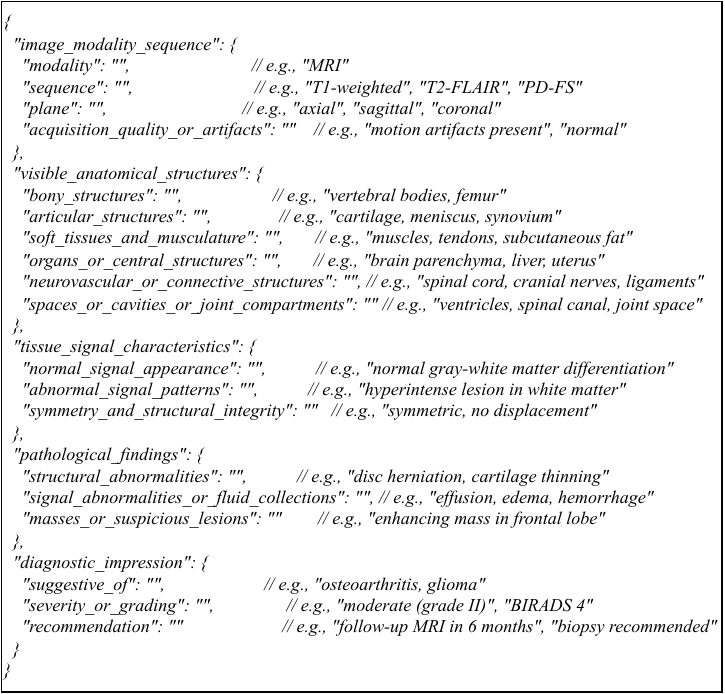}
    \caption{Hierarchical text prompt template for MRI volumes. The template defines structured fields that capture progressive levels of clinical semantics, ranging from image modality and acquisition parameters to visible anatomical structures, tissue signal characteristics, pathological findings, and diagnostic impressions.}
    \label{fig:2}
\end{figure}

To capture the textual information associated with each MRI volume, we designed a hierarchical description template that systematically organizes linguistic information from low-level imaging attributes to high-level clinical interpretations. Each description is structured across progressive levels of clinical semantics, including: (1) imaging modality and sequence parameters, (2) visible anatomical structures, (3) tissue signal characteristics, (4) pathological findings, and (5) diagnostic impression. This framework provides standardized and comprehensive text representations of MRI data, facilitating effective vision–language modeling. An example of the hierarchical description template is shown in Figure \ref{fig:2}. For majority of MRI volumes, parts of the hierarchical description can be obtained directly from existing metadata or annotations. (1) Imaging modality and sequence parameters are readily available in the DICOM header. (4) Pathological findings and (5) diagnostic impressions can often be inferred from existing ground-truth labels, bounding boxes, or structured radiology reports. In contrast, (2) anatomical structures and (3) tissue signal characteristics are rarely annotated in standard clinical datasets. These mid-level semantic descriptors require detailed anatomical recognition and nuanced radiological interpretation, making manual labeling costly and impractical at scale. 

To address this gap, we employ a generative augmentation strategy inspired by MAGPIE\cite{36_xu2024magpiealignmentdatasynthesis}. Specifically, we utilize a locally hosted Qwen-VL\cite{37_yang2025qwen3technicalreport} vision–language foundation model to automatically generate structured text descriptions directly from MRI images. This model is prompted with carefully designed instructions to produce consistent outputs for the missing components, visible anatomical structures and tissue signal characteristics. The generation process is constrained by our hierarchical template, which standardizes vocabulary and format, thereby minimizing variability in free-form outputs. Running inference on a secure local server further ensures data privacy and mitigates risks associated with cloud-based processing.

\subsubsection{Instruction-Response Data}

To enable multi-task instruction tuning within a unified framework, we constructed a set of standardized, task-specific instruction-response pairs that cast diverse MRI-related objectives into a consistent instruction-following paradigm. In this formulation, all tasks are expressed as language-guided problems, allowing a single vision–language model to learn across the full MRI workflow. Each training instance consists of three components: (1) image tokens derived from MRI volumes, (2) a natural language instruction specifying the task, and (3) the corresponding ground-truth response. Depending on the task, the output may take different forms, including reconstructed images, segmentation masks, classification labels, detection bounding, and free-text diagnostic reports.

\section{Unified Vision--Language Architecture}
To enable full-stack MRI task coverage, we design OmniMRI as a unified vision-language autoregressive Transformer. The architecture is inspired by recent advances in large-scale multimodal models such as Janus-Pro\cite{38_chen2025janusprounifiedmultimodalunderstanding}, BLIP-o3\cite{39_chen2025blip3ofamilyfullyopen}, Bagel\cite{40_deng2025emergingpropertiesunifiedmultimodal}, and Ming-Omni\cite{41_ai2025mingomniunifiedmultimodalmodel}, and uniquely adapted to volumetric MRI data and multi-task MRI settings.

\subsection{Unified Vision-Language Modeling}
The architecture of OmniMRI is illustrated in Figure \ref{fig:1}B. It is designed to jointly encode and process both visual and textual information within a unified vision–language framework, enabling the model to learn shared representations across tasks of varying granularity and complexity. Visual inputs consist of either 2D MRI slices or 3D volumetric data, which are embedded into image tokens using a Swin Transformer–based vision encoder\cite{42_Liu_2021_ICCV} that captures hierarchical spatial features across different resolutions. In parallel, textual inputs, including structured metadata from DICOM headers, descriptive clinical annotations, radiology reports, or task-specific prompts, are processed by a lightweight text tokenizer/encoder, producing language tokens. The image and language tokens are then interleaved and fused into a single sequence, which is processed by a unified autoregressive Transformer backbone adapted from Qwen2.537. This backbone employs multimodal self-attention to jointly model visual and linguistic relationships, while a mixture-of-experts feedforward network improves parameter efficiency and scalability. 

\subsection{Unified Multi-Task Decoding}
To accommodate the heterogeneity of outputs in full-stack MRI tasks, OmniMRI adopts a dual-decoder design branching from the shared vision–language Transformer backbone. An image decoder, implemented with a diffusion-based\cite{43_10081412} generation head, handles pixel-level tasks such as image reconstruction and anatomical or lesion segmentation, producing high-fidelity spatial outputs from image inputs or prompt-guided conditions. In parallel, a text decoder generates semantic-level outputs for lesion detection, diagnostic suggestion, and radiology report generation, producing structured labels or free-text descriptions in natural language. Both decoders are jointly optimized within a unified multi-task training framework, leveraging a shared cross-modal representation space. 

\section{Unified Vision-Language Training Paradigm}

To establish OmniMRI as a unified foundation model capable of addressing the full spectrum of MRI tasks, we adopt a multi-stage training pipeline that progressively builds generalizable visual representations, cross-modal reasoning ability, and instruction-following skills. The training consists of four stages. First, we pretrain the vision encoder on the image-only data in a self-supervised manner (§4.1). Second, we perform vision–language contrastive learning on paired image–text data to align visual embeddings with text embeddings (§4.2). Third, we conduct unified multimodal pretraining, where a shared autoregressive Transformer jointly models interleaved vision and language tokens to learn multimodal reasoning (§4.3). Finally, we perform multi-task instruction tuning using instruction–response pairs, allowing the model to adapt to diverse downstream tasks within a unified instruction-following framework (§4.4). 

\subsection{Vision Pretraining}

To learn robust and transferable visual representations tailored for MRI, we perform self-supervised pretraining of a vision encoder using the image-only data, without reliance on annotations. We adopt a Swin Transformer\cite{42_Liu_2021_ICCV,44_Tang_2022_CVPR,45_10.1007/978-3-031-08999-2_22} as the encoder, given its effectiveness in capturing spatial features across volumetric data. The pretraining strategy integrates multiple complementary objectives that operate at different spatial scales to capture both local anatomical details and global semantic structures. First, at the pixel level, we employ a masked image modeling task\cite{46_Xie_2022_CVPR} in which random patches of the input volume are hidden, and the model is trained to reconstruct low-level intensity distributions. Second, at the patch level, we used a contrastive objective20 that encourages representations of neighboring patches within the same volume to be more similar than those from different volumes. Third, at the volume level, we apply instance discrimination across whole MRI volumes based on acquisition characteristics such as sequence type, orientation, and anatomical content, encouraging high-level semantic grouping across data variations. Together, these objectives enable the vision encoder to capture rich anatomical priors, sequence-specific variations, and domain-invariant features, laying a strong foundation for downstream tasks ranging from low-level reconstruction to high-level semantic interpretation. 

\subsection{Vision-Language Alignment}
To align visual and textual modalities we perform vision–language contrastive learning on the paired image–text data. The pretrained vision encoder (§4.1) is frozen, and a lightweight text encoder (initialized from Qwen2.537) produces text embeddings in a shared latent space. A trainable alignment layer projects high-dimensional visual features into this space. Using a CLIP-style contrastive loss28, the model is optimized such that matched image–text pairs exhibit higher cosine similarity than mismatched ones. This stage provides a crucial bridge between unimodal visual pretraining and multimodal reasoning.

\subsection{Vision-Language Pretraining}
Building upon the aligned embedding space, we perform unified multimodal pretraining with a shared autoregressive Transformer backbone. Visual tokens (from MRI slices or volumes) and language tokens (from prompts or clinical descriptions) are interleaved into a single sequence. The model is trained with a next-token prediction objective across both modalities. Given a prefix of mixed image and text tokens, it learns to predict the next token, regardless of its modality. This joint modeling approach enables the Transformer to internalize the joint distribution of anatomy, imaging patterns, and medical language, thereby developing multimodal reasoning capabilities. The model learns to associate visual cues with semantic concepts, infer task intent, and generate clinically meaningful outputs conditioned on both visual and textual information.

\subsection{Multi-Task Instruction Tuning}

Following vision-language pretraining, we perform multi-task instruction tuning to further adapt the OmniMRI model to downstream clinical tasks using the instruction-response data. Each training instance consists of an MRI image (or volume), a natural language instruction specifying the task, and the corresponding ground-truth response. This prompt-based conditioning mechanism unifies heterogeneous MRI tasks under a single input–output format, enabling image reconstruction, image segmentation, abnormality detection, diagnostic suggestion, and report generation to be trained within the same framework. Representative instruction templates for these tasks are summarized in Table \ref{tab:task-instructions}, where the input always combines imaging data with a natural language prompt and the output corresponds to the desired task-specific response (e.g., reconstructed image, segmentation mask, bounding box, diagnostic suggestion, or free-text report.

For instance, a reconstruction task may involve a prompt like: “This is a coronal fat-suppressed MRI of the knee with 4× acceleration. Please reconstruct the image using the under-sampled data,” with the corresponding supervision being the high-resolution image. For segmentation, prompts such as “Segment the anterior cruciate ligament (ACL) in this T2-weighted axial MRI” guide the model to predict anatomical masks. Diagnostic prompts like “Does this image suggest an ACL tear?” are paired with binary or multi-label responses, while free-text generation tasks are driven by prompts such as “Generate a diagnostic impression based on this sagittal PD-FS knee MRI,” requiring the model to produce clinically coherent reports. All tasks are trained under a unified framework, using the same input-output formatting and decoding backbone, which allows the model to generalize across task types without architectural modifications. By leveraging this unified instruction–response formulation and a shared decoding backbone, OmniMRI generalizes across task types without requiring architectural modifications. Across all stages, we carefully balance data modality and task type (Table \ref{tab:training}) to maximize cross-task generalization and efficient representation reuse.

\begin{table}[t]
\centering
\caption{Representative prompt examples for full-stack MRI tasks. Each task is formulated in a unified instruction–response format, where the input consists of an MRI image (or volume) and a natural language instruction, and the output corresponds to the desired task-specific response. Examples include image reconstruction (<img> output), segmentation (<seg> mask), abnormality detection (<bbox> bounding box), diagnostic suggestion (structured text response), and report generation (free-text response).}
\renewcommand{\arraystretch}{1.35} 
\resizebox{\textwidth}{!}{ 
\begin{tabular}{ccc}
\toprule
\textbf{Task} & \textbf{Instruction Example} & \textbf{Output} \\
\midrule
Image Reconstruction & 
``\textit{\texttt{<img>} Reconstruct the image from this under-sampled MRI data.}'' &
\texttt{<img>} \\

Image Segmentation & 
``\textit{\texttt{<img>} Segment the cartilage and meniscus in this knee MRI.}'' &
\texttt{<seg>} \\

Abnormality Detection & 
``\textit{\texttt{<img>} Identify and localize any bone marrow edema or meniscus tear in this image.}'' &
\texttt{<bbox>} \\

Diagnostic Suggestion & 
``\textit{\texttt{<img>} Provide your diagnosis of this MRI volume.}'' &
\texttt{<response>} \\

Report Generation & 
``\textit{\texttt{<img>} Generate a diagnostic report based on this MRI scan.}'' &
\texttt{<response>} \\
\bottomrule
\end{tabular}}

\label{tab:task-instructions}
\end{table}

\begin{table}[!t]
\centering
\caption{Data ratio settings for each training stage. Three categories of data, image-only data, vision–text paired data, and instruction–response data, are allocated across four stages of training. Image-only data are used exclusively for vision pretraining, paired vision–text data dominate the vision–language alignment and multimodal pretraining phases, and instruction–response data are primarily employed for multi-task instruction tuning.}
\resizebox{\textwidth}{!}{%
\begin{tabular}{lcccc}
\toprule
\textbf{Dataset / Training Phase} & 
\makecell{\textbf{Vision}\\\textbf{Pretraining}} & 
\makecell{\textbf{Vision--Language}\\\textbf{Alignment}} & 
\makecell{\textbf{Vision--Language}\\\textbf{Pretraining}} & 
\makecell{\textbf{Multi-task}\\\textbf{Instruction Tuning}} \\
\midrule
Image only data            & 1   & 0   & 0   & 0 \\
Vision--Text Paired Data   & 0   & 1   & 0.8 & 0 \\
Instruction--Response Data & 0   & 0   & 0.2 & 1 \\
\bottomrule
\end{tabular}}

\label{tab:training}
\end{table}

\section{Results}
To evaluate the effectiveness of OmniMRI, we demonstrate its performance across five MRI tasks. For each task, we present several qualitative examples that highlight the model’s ability to follow diverse instructions and produce clinically meaningful outputs.

\begin{figure}[t!]
    \centering
    \includegraphics[width=0.95\textwidth]{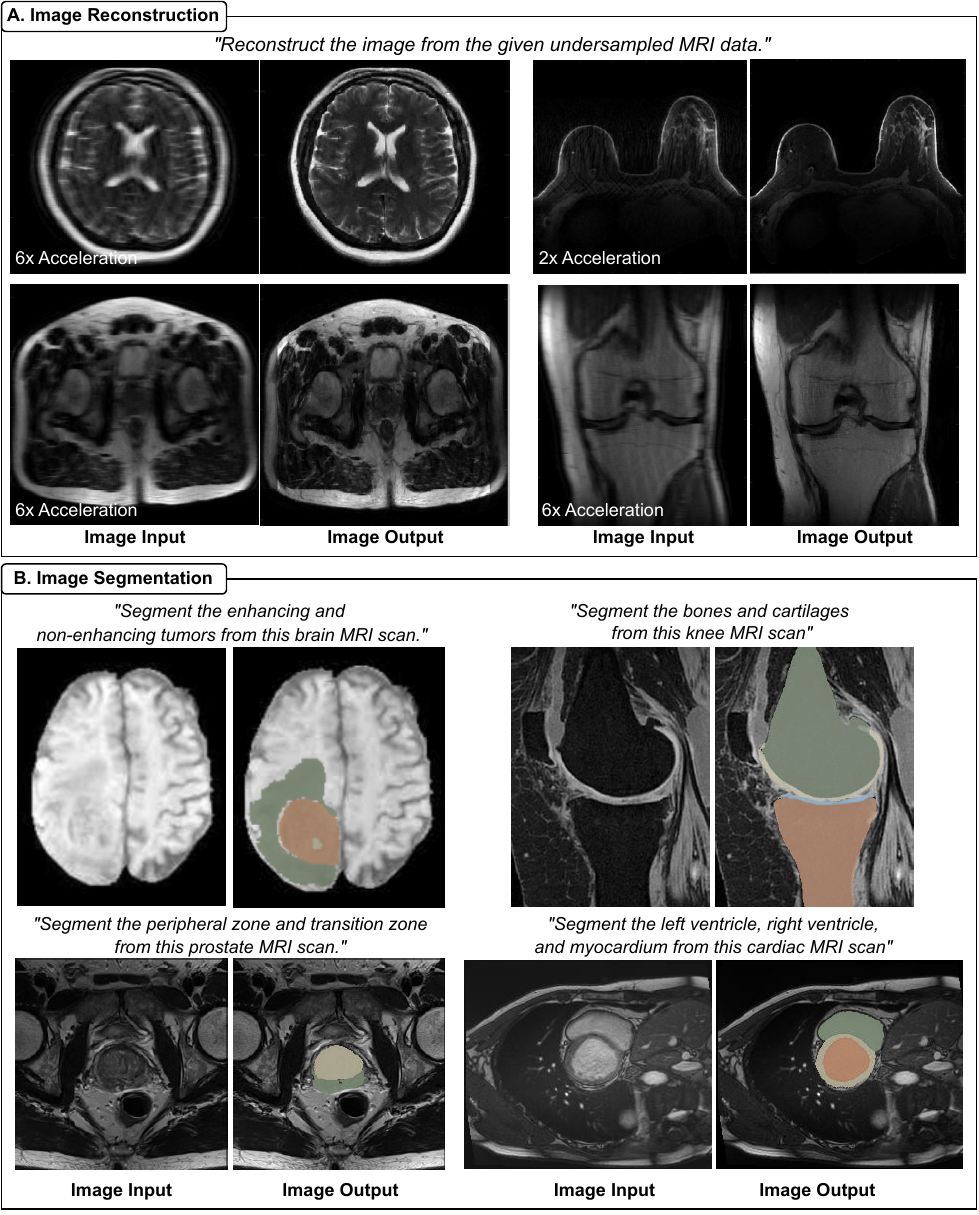}
    \caption{Qualitative results of OmniMRI on MRI reconstruction and segmentation tasks. (A) Image reconstruction. OmniMRI restores high-fidelity images from undersampled inputs across diverse anatomies, including the brain, breast, prostate, and knee. (B) Image segmentation. OmniMRI delineates anatomical and pathological. Examples include segmentation of enhancing and non-enhancing brain tumors, knee bones and cartilages, prostate peripheral and transition zones, and cardiac chambers and myocardium.}
    \label{fig:3}
\end{figure}

\textbf{MRI Reconstruction} OmniMRI demonstrates robust capability in reconstructing high-fidelity images from undersampled MRI acquisitions across diverse anatomies by following the instruction: “Reconstruct the image from the given undersampled MRI data.” As illustrated in Figure \ref{fig:3}A, the model consistently restores fine anatomical detail and suppresses undersampling artifacts across varied use cases.
\begin{itemize}
    \item Brain: Accurately reconstructs fine structural details from 6× accelerated scans, preserving cortical and subcortical features.
    \item Breast: Recovers clear tissue boundaries under 2× acceleration, supporting lesion visibility.
    \item Prostate: Maintains soft-tissue contrast even with 6× acceleration, critical for delineating glandular structures.
    \item Knee: Reconstructs sharp cartilage–bone interfaces under 6× acceleration, ensuring musculoskeletal interpretability.
\end{itemize}

Across all examples, aliasing artifacts present in the undersampled inputs are effectively suppressed, yielding images with preserved diagnostic quality and interpretability. These results highlight OmniMRI’s generalizability across anatomies, acceleration factors, and clinical use cases.

\begin{figure}[tbp]
    \centering
    \includegraphics[width=0.95\textwidth]{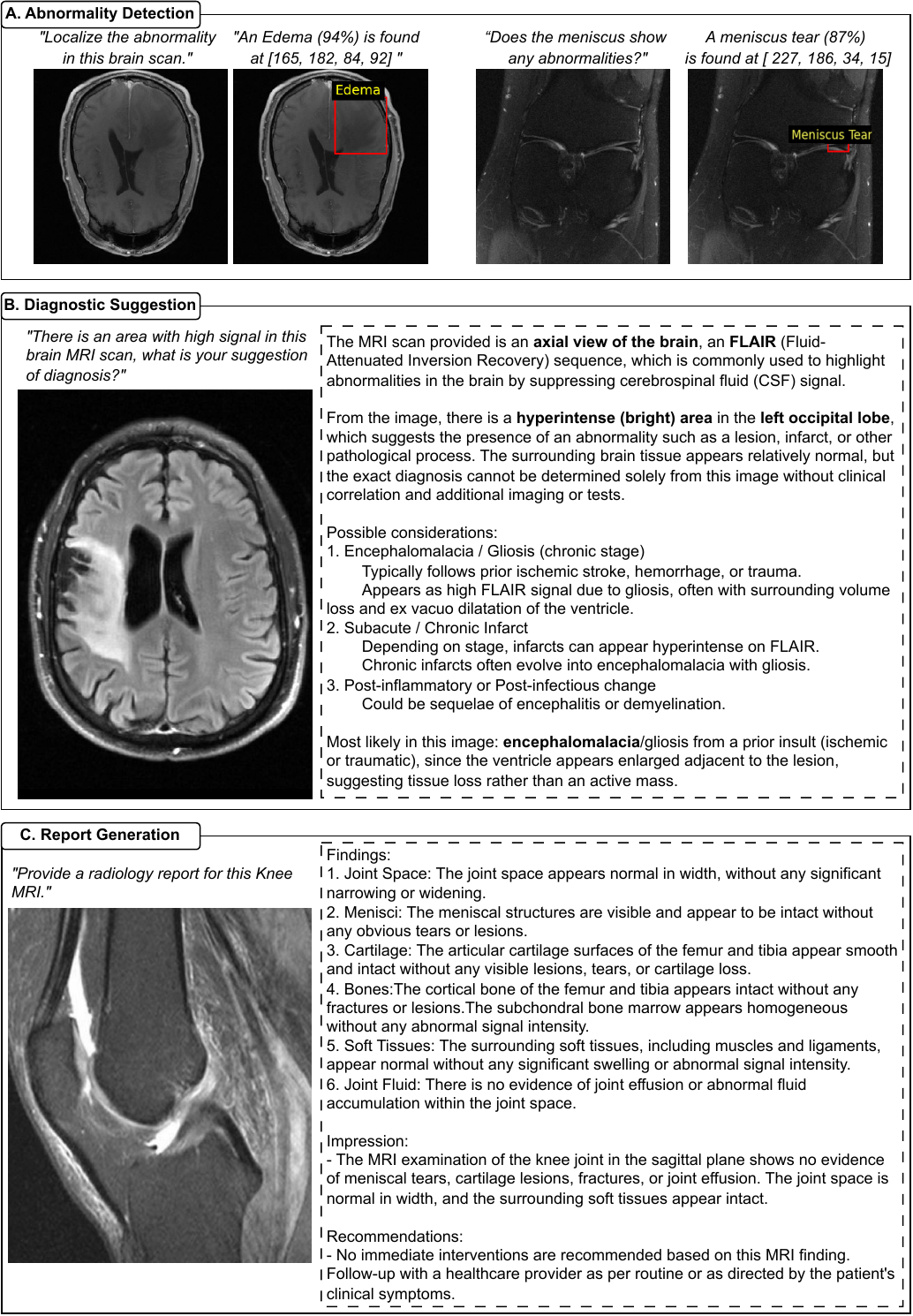}
    \caption{Qualitative results of OmniMRI on higher-level tasks. (A) Abnormality detection. The model identifies and localizes brain edema and meniscus tears in the knee, producing bounding boxes. (B) Diagnostic suggestion. OmniMRI generates structured diagnostic reasoning. In the brain MRI example, the model highlights a hyperintense region in the occipital lobe and provides differential considerations (encephalomalacia, chronic infarct, post-inflammatory change), before suggesting the most likely diagnosis. (C) Report generation. OmniMRI produces coherent radiology reports that capture anatomical context, pathological findings, impressions, and recommendations.}
    \label{fig:4}
\end{figure}

\textbf{Image Segmentation} OmniMRI demonstrates robust capability in segmenting diverse anatomical structures and pathological regions across multiple organ systems by following organ-specific instructions. As illustrated in Figure \ref{fig:3}B, the model accurately delineates:
\begin{itemize}
    \item Brain: Enhancing and non-enhancing tumor regions with precise spatial boundaries.
    \item Knee: Bony structures and cartilaginous tissue, critical for musculoskeletal assessment.
    \item Prostate: Peripheral and transition zones, supporting cancer localization.
    \item Heart: Cardiac chambers and myocardium with well-defined borders.
Across these examples, the predicted masks exhibit sharp boundary definition and anatomical fidelity, maintaining alignment with clinical landmarks. 
\end{itemize}

The consistency of segmentation performance across heterogeneous organs and tissue types of underscores OmniMRI’s ability to generalize segmentation instructions, bridging applications from neuroimaging to musculoskeletal, body, and cardiac MRI.

\textbf{Abnormality Detection} OmniMRI demonstrates the capability to detect and localize abnormalities across diverse anatomical regions by generating bounding boxes and region-level predictions. As shown in Figure \ref{fig:4}A, the model accurately identifies edema in the brain, producing bounding boxes with high confidence scores that align closely with radiologically confirmed pathological regions. Notably, OmniMRI also captures subtle abnormalities such as meniscal tears in the knee, which are often visually challenging to detect. These results highlight the model’s ability to perform both coarse- and fine-grained abnormality detection across different organ systems, supporting diagnostic workflows that require sensitivity to subtle imaging findings while maintaining robustness across broader pathologies.

\textbf{Diagnostic Suggestion} When prompted with diagnostic queries, OmniMRI produces structured, clinically oriented suggestions that integrate imaging observations with medical reasoning. As illustrated in the brain MRI example (Figure \ref{fig:4}B), the model identifies a hyperintense region in the left occipital lobe and provides a differential diagnosis that includes encephalomalacia, chronic infarct, and post-inflammatory changes, before suggesting the most likely interpretation. This output demonstrates OmniMRI’s capacity to move beyond binary classification, instead offering interpretive reasoning that mirrors radiological practice. By synthesizing imaging features with clinical knowledge, OmniMRI enables hypothesis generation and decision support, highlighting its potential role as an assistive tool in radiology workflows and multidisciplinary care.

\textbf{Report Generation} OmniMRI is capable of generating coherent, structured, and clinically meaningful radiology reports directly from MRI images and task-specific instructions. In the knee MRI example, the model-produced report systematically addresses each relevant component of interpretation: Joint space evaluation; Menisci morphology and integrity; Cartilage thickness and defects; Bone structures and marrow signal; Surrounding soft tissues. The report then concludes with a diagnostic impression and actionable recommendations, mirroring the organization and language of expert radiology practice. By providing both anatomical context and pathology assessment in natural language format, OmniMRI outputs demonstrate not only interpretive accuracy but also workflow relevance. These results highlight the model’s potential to unify image analysis and structured reporting, thereby reducing documentation burden for radiologists while maintaining clinical fidelity and consistency.

\section{Conclusion}
In this work, we introduced OmniMRI, a unified vision–language foundation model designed to span the full MRI workflow—from low-level tasks such as image reconstruction and segmentation to higher-level functions including abnormality detection, diagnostic suggestion, and radiology report generation. Unlike prior approaches narrowly tailored to specific anatomies, contrasts, or isolated tasks, OmniMRI leverages a single multimodal architecture with instruction-following capability, enabling flexible adaptation across diverse clinical settings.

By integrating a large-scale, heterogeneous MRI corpus with a multi-stage training paradigm, comprising self-supervised vision pretraining, vision–language alignment, multimodal pretraining, and multi-task instruction tuning—OmniMRI learns rich, transferable representations that support both zero-shot generalization and supervised specialization. Our qualitative evaluations demonstrate its ability to:
\begin{itemize}
    \item Restore high-fidelity reconstructions from undersampled acquisitions,
    \item Generate accurate anatomical and pathological segmentations,
    \item Localize subtle abnormalities across organs,
    \item Provide structured diagnostic reasoning, and
    \item Produce clinically coherent radiology reports.
\end{itemize}

Together, these results highlight a pathway toward generalist foundation models in medical imaging, showing that traditionally fragmented, task-specific pipelines can be consolidated into a scalable framework that unifies image understanding with language-based clinical reasoning.

While our current evaluation focuses on qualitative demonstrations, future directions include large-scale quantitative benchmarking, prospective validation with radiologist readers, and exploration of deployment strategies for integration into clinical workflows. These steps will be essential to fully realize OmniMRI’s potential to augment radiology practice, reduce workflow burden, and advance precision imaging.

\bibliographystyle{unsrt}
\bibliography{OmniMRI_refs}

\end{document}